%% file: main.tex
\documentclass[10pt,twocolumn,letterpaper]{article}

\usepackage[pagenumbers]{cvpr} 

\input{preamble}

\definecolor{cvprblue}{rgb}{0.21,0.49,0.74}
\usepackage[pagebackref,breaklinks,colorlinks,citecolor=cvprblue]{hyperref}

\usepackage{bm}
\usepackage{multirow}
\usepackage{bbding}
\usepackage{color}
\usepackage[linesnumbered,ruled,vlined]{algorithm2e}

\def\paperID{5542} 
\def\confName{CVPR}
\def\confYear{2024}

\title{I\&S-ViT: An Inclusive \& Stable Method for Pushing the Limit of Post-Training ViTs Quantization} 

\author{
Yunshan Zhong$^{1,2}$, Jiawei Hu$^2$, Mingbao Lin$^3$, Mengzhao Chen$^2$, Rongrong Ji$^{1,2,4}$\thanks{Corresponding Author: rrji@xmu.edu.cn}\\
$^1$Institute of Artificial Intelligence, Xiamen University\\
$^2$MAC Lab, Department of Artificial Intelligence, School of Informatics, Xiamen University \, \\ $^3$ Rakuten $^4$Peng Cheng Laboratory,  \\
{\tt\small zhongyunshan@stu.xmu.edu.cn, jiaweihu@stu.xmu.edu.cn} \\ 
{\tt\small linmb001@outlook.com, cmzxmu@stu.xmu.edu.cn, rrji@xmu.edu.cn}
}

\begin{document}
\maketitle

\begin{abstract}

%

Albeit the scalable performance of vision transformers (ViTs), the dense computational costs (training \& inference) undermine their position in industrial applications.
%
%
%
Post-training quantization (PTQ), tuning ViTs with a tiny dataset and running in a low-bit format, well addresses the cost issue but unluckily bears more performance drops in lower-bit cases.
%
%
%
%
%
In this paper, we introduce I\&S-ViT, a novel method that regulates the PTQ of ViTs in an inclusive and stable fashion.
I\&S-ViT first identifies two issues in the PTQ of ViTs:
(1) Quantization inefficiency in the prevalent log2 quantizer for post-Softmax activations;
(2) Rugged and magnified loss landscape in coarse-grained quantization granularity for post-LayerNorm activations.
Then, I\&S-ViT addresses these issues by introducing:
(1) A novel shift-uniform-log2 quantizer (SULQ) that incorporates a shift mechanism followed by uniform quantization to achieve both an inclusive domain representation and accurate distribution approximation;
(2) A three-stage smooth optimization strategy (SOS) that amalgamates the strengths of channel-wise and layer-wise quantization to enable stable learning.
Comprehensive evaluations across diverse vision tasks validate I\&S-ViT's superiority over existing PTQ of ViTs methods, particularly in low-bit scenarios. For instance, I\&S-ViT elevates the performance of 3-bit ViT-B by an impressive 50.68\%. Code: \url{https://github.com/zysxmu/IaS-ViT}.

\end{abstract}
\section{Introduction}


%
In the ever-evolving realm of computer vision, vision transformers (ViTs) of late~\cite{DosovitskiyZ21An} stand out as an excellent architecture to capture the long-range relationships among image patches with multi-head self-attention (MHSA) mechanism. 
%
%
%
%
%
However, exceptional power comes at the expense of great computing: $n$ image patches result in $\mathcal{O}(n^2)$ complexity from the MHSA operation. In order to provide affordable usage of ViTs, researchers from the vision community have strained every nerve to reduce the compute costs~\cite{li2023repq,bolya2022token,lin2022fqvit,lin2023super,chen2023diffrate}.

%
%

Model quantization reduces the representation precision of weights \& activations, and has garnered sustainable attention due mostly to its reliable academic support and applied industrial practice~\cite{whitepaper}.
A multitude of studies~\cite{li2023vit,Yang2023Oscillation,li2022q,li2022qv,zhong2023multiquant,gong2019differentiable,LSQ,zhong2022dynamic} have run into quantization-aware training (QAT) by accessing the entire training dataset and executing an end-to-end retraining. Such premises require a very dense computational cost in network retraining, which sadly drops an obstacle to the broad deployment of QAT methods.
%
%
%
%
Therefore, researchers have gravitated to post-training quantization (PTQ) in search of quantizing models with a tiny dataset, for the sake of minor costs~\cite{ACIQ,li2022patch,lin2022fqvit,liu2023noisyquant,frumkin2023jumping}.
To adapt to the specific structure in ViTs such as LayerNorm and self-attention mechanisms, current efforts on PTQ of ViTs typically introduce dedicated quantizers and quantization schemes to maintain ViTs' original performance.
To adapt to the unique components in ViTs such as LayerNorm and self-attention operations, these efforts introduce dedicated quantizers and schematic quantization to maintain ViTs' performance.
For example, FQ-ViT ~\cite{lin2022fqvit} and PTQ4ViT~\cite{yuan2022ptq4vit} respectively introduce a log2 quantizer and a twin uniform quantizer for post-Softmax activations. RepQ-ViT~\cite{li2023repq} adopts the channel-wise quantizer for high variant post-LayerNorm activations first and then reparameterizes it to a layer-wise quantizer. 
%
%
Notwithstanding, considerable performance drops are observed when performing low-bit quantization. By way of illustration, in 4-bit, RepQ-ViT~\cite{li2023repq} causes 10.82\% accuracy drops over full-precision DeiT-S~\cite{touvron2021training} on ImageNet~\cite{russakovsky2015imagenet}; while in 3-bit, it leads to 74.48\% accuracy drops.
Recent optimization-based PTQ methods have demonstrated their capacity in quantizing convolutional neural networks (CNNs)~\cite{li2021brecq,wei2021qdrop,liu2023pd}.
However, their attempts in ViTs remain unexploited, and in Tab.\,\ref{tab:imagenet-0} of this paper we find their applications typically result in overfitting in high-bit cases and suffer large performance degradation in ultra-low bit cases, which in turn, barricades their capacity in ViTs architectures~\cite{ding2022towards,lin2023bit,li2023repq,liu2021post}.

%

%

In this paper, we present a novel optimized-based PTQ method specifically tailored for ViTs, called I\&S-ViT, to harness the potential of optimized-based techniques.
At first, we identify that the log2 quantizer, widely adopted for long-tailed post-Softmax activations, suffers from the quantization inefficiency issue which refers to the representative range failing to encompass the entire input domain. In response, we propose a shift-uniform-log2 quantizer (SULQ). This novel quantizer, by introducing an initial shift bias to the log2 function input, subsequently uniformly quantizes its outputs. SULQ is able to fully include the input domain to solve the quantization inefficiency issue and accurately approximate the distribution of post-Softmax activations. Moreover, SULQ can be efficiently executed by the fast and hardware-friendly bit-shifting operations~\cite{li2023repq,lin2022fqvit}.


%
%
%

Furthermore, we observe marked distinctions in the loss of landscapes across different quantization granularity.
As shown in Fig.\,\ref{fig:loss-landscape}, Channel-wise weight quantization and layer-wise post-LayerNorm activation quantization result in a rugged and magnified loss, impeding quantization learning and compromising model performance~\cite{bai2021binarybert,huang2022sdq,frumkin2023jumping}.
This aggravation can be alleviated if executing full-precision weights.
Further applying channel-wise quantization to post-LayerNorm activations results in a smooth landscape with reduced loss magnitudes, leading to more stable and effective optimization~\cite{li2018visualizing,lin2023bit}.
Motivated by these insights, we propose a three-stage smooth optimization strategy (SOS) to harness the benefits of the smooth and low-magnitude loss landscape for optimization, while maintaining the efficiency of the layer-wise quantization for activations~\cite{li2023repq,BitSplitStitching,whitepaper}. 
In the first stage, we fine-tune the model with full-precision weights alongside channel-wise quantized post-LayerNorm activations, and other activations employ a layer-wise quantizer. 
In the second stage, we seamlessly transit the channel-wise quantizer to its layer-wise counterpart with the scale reparameterization technique~\cite{li2023repq}.
Finally, in the third stage, the model undergoes fine-tuning with both activations and weights subjected to quantization for restoring the performance degradation of weights quantization.

Comprehensive experimental assessments across a wide range of ViT variants and vision tasks validate the preeminence of the proposed I\&S-ViT. 
For instance, for the 3-bit ViT-B, I\&S-ViT significantly elevates performance, registering an encouraging improvement of 50.68\%.

\section{Related Work}

\subsection{Vision Transformers (ViTs)}
%
%
%
%

Subsequent to CNNs, ViTs~\cite{DosovitskiyZ21An} have again revolutionized the field of computer vision.
ViTs tokenize an image as the input of a transformer architecture~\cite{vaswani2017attention}, therefore a structured image is processed in a sequence fashion.
Given that the performance of vanilla ViTs relies on the large-scale pre-trained dataset, DeiT~\cite{touvron2021training} develops an efficient teacher-student training approach.
In addition to image classification, ViTs have been well adopted in low-lever vision~\cite{liang2021swinir} and video process~\cite{arnab2021vivit}, \emph{etc}.
Liang~\emph{et al}.~\cite{liang2021swinir} proposed SwinIR that builds on Swin transformers block to solve image restoration tasks.
In~\cite{arnab2021vivit}, a pure-transformer model is proposed for video classification, wherein spatio-temporal tokens from videos are encoded using a series of transformer layers.
In particular, Swin's hierarchical structure with the shifted window-based self-attention~\cite{liu2021swin}, extends ViTs' applicability to dense vision tasks such as object detection~\cite{carion2020end,zhu2020deformable} and segmentation~\cite{zheng2021rethinking}.
However, the impressive performance of ViTs relies on significant computational overhead, making them challenging for resource-constrained environments~\cite{mehta2021mobilevit,zhang2022minivit}.

\subsection{ViTs Quantization}
%
%
%
By reducing the numerical precision, model quantization has been instrumental in providing deployment for neural networks~\cite{whitepaper}. 
%
%
Despite the efficacy of quantization-aware training (QAT) in retraining performance, its deficiency includes accessibility to the complete training set and the nature of compute-heavy retraining~\cite{gong2019differentiable,LSQ,zhong2022exploiting,zhong2022intraq}.
%
%
%
Therefore, the research pivot has shifted to post-training quantization (PTQ) for ViTs, with its small dataset requirement and fast industrial deployment~\cite{zhong2023distribution,ACIQ,Upordown,zhong2022fine}. 
Unfortunately, the customized operators, such as LayerNorm and MHSA in ViTs, create maladjustments when making a direct extension of PTQ methods from CNNs to ViTs~\cite{li2021brecq,wei2021qdrop,li2023repq,lin2022fqvit}.

Consequently, there is a growing consensus to develop ViTs-specialized PTQ methods.
%
FQ-ViT~\cite{lin2022fqvit} introduces a fully-quantized method for ViTs, incorporating Powers-of-Two Scale and Log-Int-Softmax for LayerNorm and post-Softmax activations. Liu \emph{et al}.~\cite{liu2021post} embedded a ranking loss into the quantization objective to maintain the relative order of the post-Softmax activations, combined with a nuclear norm-based mixed-precision scheme. 
PTQ4ViT~\cite{yuan2022ptq4vit} adopts a twin uniform quantization method to reduce the quantization error on activation values, complemented by a Hessian-guided metric for searching quantization scales.
Liu \emph{et al}.~\cite{liu2023noisyquant} suggested adding a uniform noisy bias to activations. 
APQ-ViT~\cite{ding2022towards} establishes a calibration strategy that considers the block-wise quantization error.
Evol-Q~\cite{frumkin2023jumping} adopted an evolutionary search to determine the disturbance-sensitive quantization scales.
~\cite{lin2023bit} proposed gradually decreasing the bit-width to achieve a good initialization point.
RepQ-ViT~\cite{li2023repq} first deploys complex quantizers for post-LayerNorm activations, subsequently simplifying these quantizers through reparameterization.
%

\begin{figure*}[!ht]
\centering
\subfloat[
{\centering \label{fig:lq-com}}]
{
\centering\includegraphics[height=0.2\linewidth]{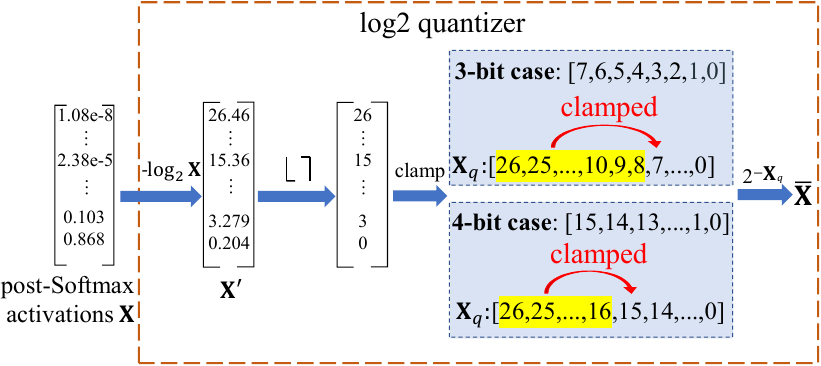} \hfill
}
\subfloat[
{\centering  \label{fig:sulq-com}}]
{
\centering\includegraphics[height=0.2\linewidth]{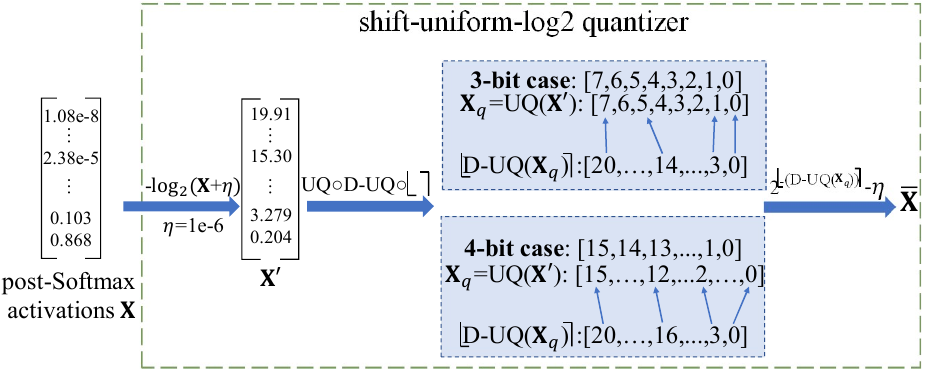} \hfill
}
\caption{Illustration of (a) the quantization inefficiency issue of the 3/4-bit log2 quantizers. (b) the quantization process of 3/4-bit shift-uniform-log2 quantizers.}
\label{fig:Q-com}
\end{figure*}

\section{Preliminaries}

\textbf{Structure of ViTs}. An input image $I$ is first split into $N$ flattened 2D patches, which are then projected by an embedding layer to $D$-dimensional vectors, denoted as $\mathbf{X}_0\in \mathbb{R}^{N\times D}$. Then, $\mathbf{X}_0$ is fed into $L$ transformer blocks, each of which consists of a multi-head self-attention (MHSA) module and a multi-layer perceptron (MLP) module. For the $l$-th transformer blocks, the computation can be expressed as:
\begin{align}
  \mathbf{Z}_{l-1} & = \text{MHSA}_{l}(\text{LayerNorm}(\mathbf{X}_{l-1})) + \mathbf{X}_{l-1}. \\
  \mathbf{X}_l & = \text{MLP}_{l}(\text{LayerNorm}(\mathbf{Z}_{l-1})) + \mathbf{Z}_{l-1}.
\end{align}

MHSA consists of $H$ self-attention heads. For the $h$-th head, the operations with input $\mathbf{X}_{l-1, h}$ formulated below:
%
%
\begin{align}
    [\mathbf{Q}_h, \mathbf{K}_h, \mathbf{V}_h] & = \mathbf{X}_{l-1, h} \bm{W}^{QKV}_h  + \bm{b}^{QKV}_h. \\
    \mathbf{A}_{h} &= \text{Softmax}\left(\frac{\mathbf{Q}_h\cdot \mathbf{K}_h^T}{\sqrt{D_h}}\right)\mathbf{V}_h,
\end{align}
where $D_h$ is the dimension size of each head. Denoting $\mathbf{X}_{l-1} = concat(\mathbf{X}_{l-1, 1}, \mathbf{X}_{l-1, 2},...,\mathbf{X}_{l-1, H})$, the results of each head are concatenated and the output of the $l$-th MHSA is obtained by:
\begin{align}
    \text{MHSA}(\mathbf{X}_{l-1}) = concat(\mathbf{A}_{1},\mathbf{A}_{2},\ldots, \mathbf{A}_{H}) \bm{W}^O + \bm{b}^O.
\end{align}

The MLP module contains two fully-connected layers (FC) and the GELU activation function. Denoting the input to the $l$-th MLP module as $\mathbf{Z}_{l-1}$, the calculation is as:
\begin{equation}
  \text{MLP}(\mathbf{Z}_{l-1}) = \text{GELU}(\mathbf{Z}_{l-1} \bm{W}^1+\bm{b}^1)\bm{W}^2 + \bm{b}^2.
\end{equation}

It can be seen that the major computation costs of ViTs come from the large matrix multiplications. Therefore, as a common practice in previous works~\cite{li2023repq,yuan2022ptq4vit}, we choose to quantize all the weights and inputs of matrix multiplications, leaving LayerNorm and Softmax operations as full-precision types.

\textbf{Quantizers}. 
The uniform quantizer evenly maps full-precision values $\mathbf{X}$ to integer $\mathbf{X}_q$. Given bit-width $b$, the uniform quantizer (UQ) is formally defined as:
\begin{align}
\label{eq:UQ}
  \mathbf{X}_q = \text{UQ}(\mathbf{X}, b) = \text{clamp}\left(\left\lfloor \frac{\mathbf{X}}{s} \right\rceil+z, 0, 2^b-1 \right),
\end{align}
where $\left\lfloor\cdot\right\rceil$ denotes the round function, clamp constrains the output between $0$ and $2^b-1$, $s$ and $z$ respectively are the quantization scale and the zero-point:
\begin{equation}
\label{eq:sz}
s = \frac{\max(\mathbf{X})-\min(\mathbf{X})}{2^b-1}, \quad z = \left\lfloor-\frac{\min(\mathbf{X})}{s} \right\rceil.
\end{equation}

Then, the de-quantized values $\Bar{\mathbf{X}}$ can be calculated with de-quantization process D-UQ:
\begin{align}
\label{eq:D-UQ}
  \Bar{\mathbf{X}} = \text{D-UQ}(\mathbf{X}_q) = s\left(\mathbf{X}_q-z\right) \approx \mathbf{X}.
\end{align}

To handle the nature of the long-tail distribution of post-Softmax activations, the log2-based quantizer~\cite{cai2018deep} has been extensively adopted in many previous PTQ methods of ViTs~\cite{li2023repq,lin2022fqvit,frumkin2023jumping}.
A common choice is using the log2 quantizer (\text{LQ}) for non-negative post-Softmax activation $\mathbf{X}$:
\begin{align}
 \mathbf{X}_q = \text{LQ}(\mathbf{X}, b) = \text{clamp}\left(\left\lfloor -\log_2 \frac{\mathbf{X}}{s} \right\rceil, 0, 2^b-1 \right).
\end{align}

Then, the de-quantization process D-LQ is used to obtain de-quantized values $\Bar{\mathbf{X}}$:
\begin{align}
  \Bar{\mathbf{X}} = \text{D-LQ}(\mathbf{X}_q) = s \cdot 2^{-\mathbf{X}_q} \approx \mathbf{X}.
\end{align}

For consistency with earlier works~\cite{yuan2022ptq4vit,li2023repq,ding2022towards}, we utilize the channel-wise quantizer for weights and the layer-wise quantizer for activations.


\begin{figure*}[!ht]
\centering
\subfloat[
{\centering \label{fig:Q-funtion-a}}]
{
\centering\includegraphics[height=0.23\linewidth]{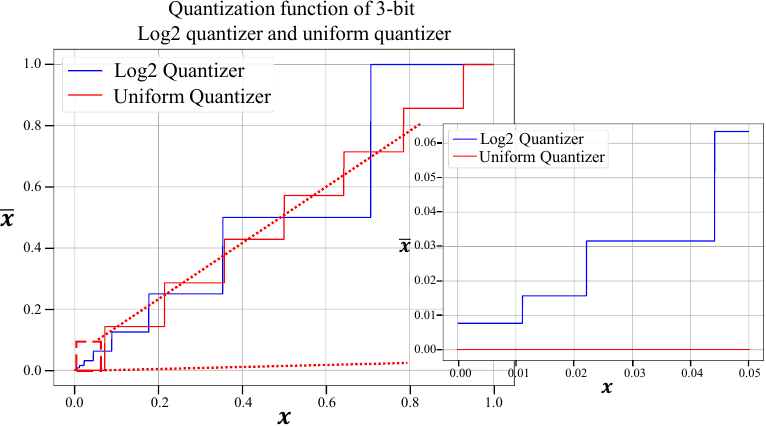} \hfill
}
\hspace{10mm}
\subfloat[
{\centering  \label{fig:Q-funtion-b}}]
{
\centering\includegraphics[height=0.23\linewidth]{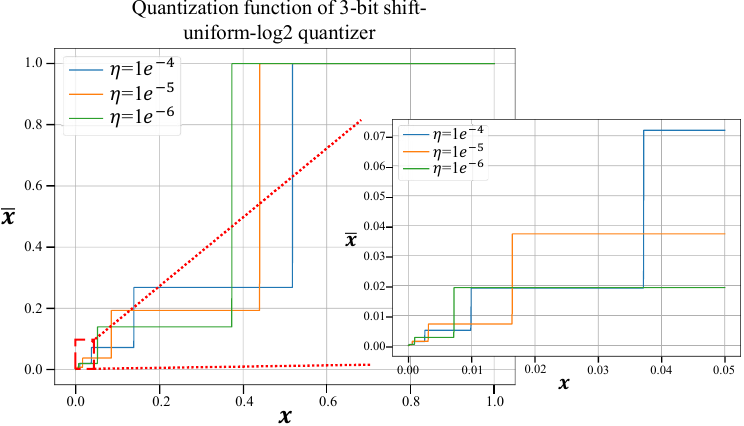} \hfill
}
\caption{Illustration of the quantization function of 3-bit (a) log2 quantizer and uniform quantizer. (b) shift-uniform-log2 quantizer.}
\label{fig:Q-function}
\end{figure*}

\section{Method}

\subsection{Block-wise Optimization}

In alignment with \cite{li2021brecq,wei2021qdrop,ding2022towards}, we establish the block-wise reconstruction as the learning objective. Let $\mathbf{X}_l$ represent outputs of the $l$-th full-precision transformer block, and $\Bar{\mathbf{X}}_l$ represent outputs of the quantized version. The block-wise reconstruction is defined as:
\begin{align}
  \mathcal{L}_l = \| \mathbf{X}_l - \Bar{\mathbf{X}}_l \|_2.
\end{align}

Note that $\mathcal{L}_l$ is only backward to update weights in the $l$-th transformer block.
In the next, we delve into the challenges and corresponding solutions. In Sec.\,\ref{sec:sulq}, we first identify the quantization inefficiency issue of log2 quantizer, and thus introduce our solution, \emph{i.e.}, shift-uniform-log2 quantizer. In Sec.\,\ref{sec:sos-obs}, we find that scale smoothness varies across different quantization granularity, and thus propose our solution, \emph{i.e.}, smooth optimization strategy.


\subsection{Shift-Uniform-Log2 Quantizer}
\label{sec:sulq}
In Fig.\,\ref{fig:Q-funtion-a}, we plot the relationship of full-precision $\mathbf{X}$ and de-quantized $\bar{\mathbf{X}}$ when uniform quantizer and log2 quantizer are deployed.
Compared to the uniform quantizer, the log2 quantizer prioritizes more bits for the near-zero region, showing its advantage in addressing the prevalent long-tail distribution in post-Softmax activations~\cite{dong2023packqvit,li2023repq,lin2022fqvit,frumkin2023jumping}. \textit{However, log2 quantizer, as we analyze below, also exhibits a primary issue of quantization inefficiency}.


%

In Fig.\,\ref{fig:lq-com}, we give an example to elucidate what the issue is.
Considering the input post-Softmax activations $\mathbf{X}$ with a range of [1.08-8, 0.868], the rounded results have a span of maximal 26 and minimal 0. The 3-bit quantization covers a range of [0, 7], therefore, the rounded segment [8, 26] would be clamped to 7. As for 4-bit quantization, the rounded segment [16, 26] would be clamped to 15.
We name it ``quantization inefficiency'' in that a large portion of values are clamped to a position at remote.
The post-Softmax activations have a plethora of zero-around values. The quantization inefficiency issue causes large quantization errors and compromises the model's performance.

%
%

%
Inspired by the above analyses, we introduce the shift-uniform-log2 quantizer (SULQ) to address the quantization inefficiency issue. In particular, we first include a shift bias $\eta$ before feeding the full-precision input $\mathbf{X}$ to the log2 transformation, and then follow a uniform quantizer.
%
%
\begin{align}
 \mathbf{X}_q = \text{SULQ}(\mathbf{X}, b) =  \text{UQ}\left( -\log_2 (\mathbf{X}+\eta), b \right). 
\end{align}

The de-quantization process of our SULQ is derived as:
\begin{align}
  \Bar{\mathbf{X}} = \text{D-SULQ}(\mathbf{X}_q) = 2^{ \left\lfloor-(\text{D-UQ}(\mathbf{X}_q))\right\rceil}-\eta \approx \mathbf{X}.
\end{align}

The ``UQ'' and ``D-UQ'' respectively denote the uniform quantizer in Eq.\,(\ref{eq:UQ}) and the corresponding de-quantization process in Eq.\,(\ref{eq:D-UQ}). Note that the round function $\left\lfloor\cdot\right\rceil$ is applied to the outputs of $\text{D-UQ}(\mathbf{X}_q)$ to ensure integer outputs, such that fast and hardware-friendly bit-shifting operations can be applied~\cite{li2023repq,lin2022fqvit}.
Fig.\,\ref{fig:Q-funtion-b} presents the relationship of full-precision $\mathbf{X}$ and de-quantized $\bar{\mathbf{X}}$, \emph{w.r.t.} different $\eta$, for ease of comparison with the uniform quantizer and log2 quantizer in Fig.\,\ref{fig:Q-funtion-a}.
Also, Fig.\,\ref{fig:sulq-com} presents the 3/4-bit quantization processes of our SULQ.
The proposed SULQ enjoys two advantages:

First, our SULQ well solves the quantization inefficiency issue of the log2 quantizer.
In particular, by leveraging the uniform quantizer, SULQ inclusively represents the full range of the input domain. As showcased in Fig.\,\ref{fig:sulq-com}, for the 3-bit case, SULQ uniformly allocates the 8 integers across the range of input values. Consequently, the output of $\left\lfloor-(\text{D-UQ}(\mathbf{X}_q))\right\rceil$ uniformly spans the range of [19, 0].
Similarly, for the 4-bit case, all 16 integers are employed to uniformly include the range of [19, 0].
This design ensures that SULQ accurately retains the near-zero values. For example, for the 3-bit case, given the input value of 2.38e-5, SULQ quantizes it to 6.00e-5, while the log2 quantizer quantizes it to 7.81e-3. Clearly, SULQ yields a smaller quantization error.

Second, as shown in Fig.\,\ref{fig:Q-funtion-b}, SULQ employs a fine-grained quantization bit allocation strategy for regions proximate to zero while allocating sparser bits for areas near one. This allocation paradigm well matches the long-tail distribution of post-Softmax activations.
Additionally, Fig.\,\ref{fig:Q-funtion-b} reveals that varying the parameter $\eta$ leads to disparate quantization point distributions. Consequently, by adjusting $\eta$, SULQ can adapt to diverse input distributions. This introduces a higher flexibility than the log2 quantizer, whose quantization points are only distributed in a fixed pattern. 

\begin{figure*}[!t]
\centering
\subfloat[
{\centering  \label{fig:w-q-a-lq}}]
{
\centering\includegraphics[width=0.25\linewidth]{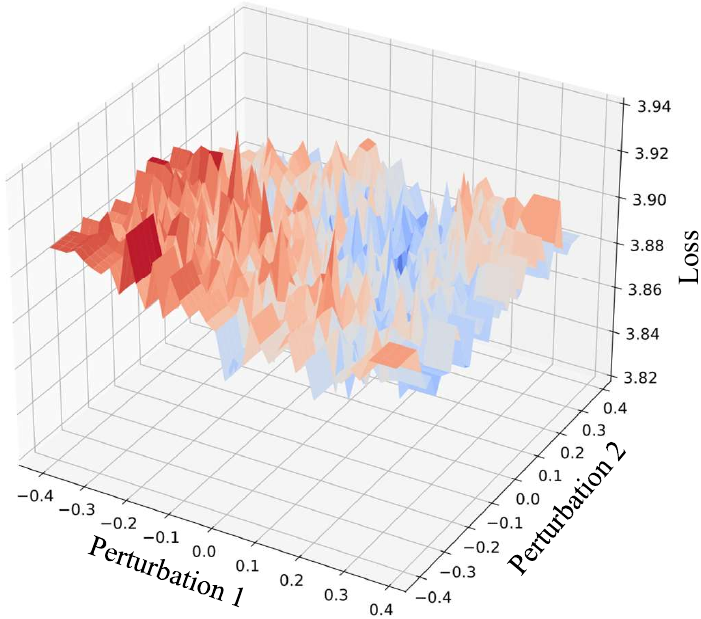} \hfill
}
\hspace{8mm}
\subfloat[
{\centering  \label{fig:w-fp-a-lq}}]
{
\centering\includegraphics[width=0.25\linewidth]{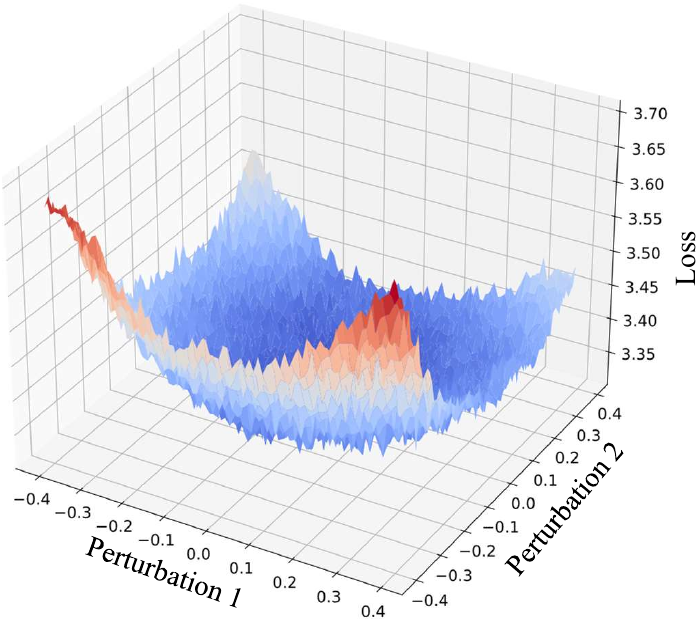} \hfill
}
\hspace{8mm}
\subfloat[
{\centering \label{fig:w-fp-a-cq}}]
{
\centering\includegraphics[width=0.25\linewidth]{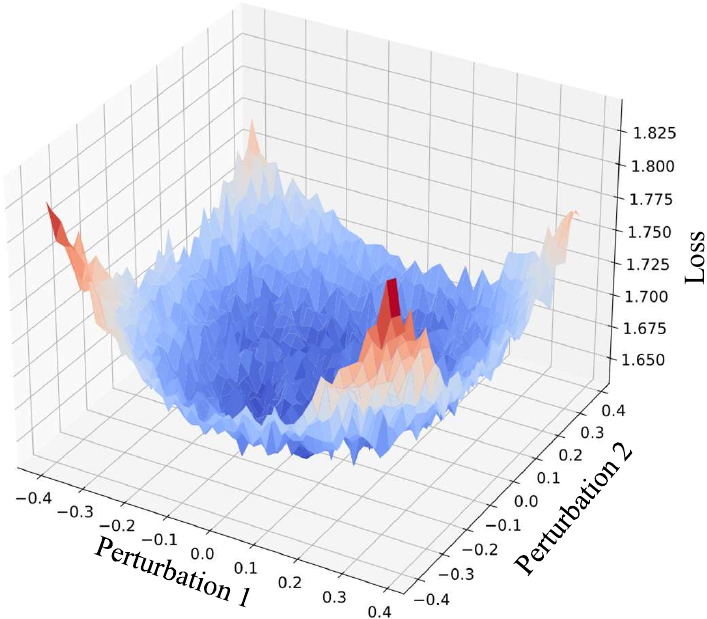} \hfill
}
\caption{Loss landscapes for the 4-bit DeiT-S in transformer block 10. We perturb the weights along two basis vectors (Perturbation 1 \& 2) to visualize the loss landscape. (a) Channel-wise weight quantization \& layer-wise activation quantization. (b) Full-precision weights \& layer-wise activation quantization. (c) Full-precision weights \& channel-wise activation quantization.}
\label{fig:loss-landscape}
\end{figure*}

Compared with the log2 quantizer, SULQ only involves one extra round function and two addition operations, the costs of which are negligible.
During the inference, SULQ produces integer outputs. As a result, its computations can be efficiently executed by fast and hardware-friendly bit-shifting operations, in line with previous works~\cite{li2023repq,lin2022fqvit,frumkin2023jumping}.
It is worth noting that many preceding methods perform transformations before executing uniform quantization, such as normalization~\cite{APoT,qin2020forward,xu2021recu} and power functions~\cite{jung2019learning,yvinec2022powerquant}. However, these methods focus on weight quantization. In contrast, our SULQ is specifically tailored for post-Softmax activations by addressing the observed quantization inefficiency issue in the log2 quantizer, which remains largely untapped in prior research.

\begin{table*}[ht]
\centering
\small
\begin{tabular}{cccccccccc}
\toprule[1.25pt]
\textbf{Method} & \textbf{Opti.} &\textbf{Bit. (W/A)} & \textbf{ViT-S}& \textbf{ViT-B} & \textbf{DeiT-T}& \textbf{DeiT-S} & \textbf{DeiT-B} &  \textbf{Swin-S}& \textbf{Swin-B} \\
\midrule[0.75pt]
Full-Precision & -  & 32/32 & 81.39 & 84.54 & 72.21 & 79.85 & 81.80 & 83.23 & 85.27 \\ 
\midrule[0.75pt]
PTQ4ViT~\cite{yuan2022ptq4vit} & $\times$ & 3/3 & 0.01 & 0.01 & 0.04 & 0.01 & 0.27 & 0.35 & 0.29 \\
BRECQ~\cite{li2021brecq} &  $\checkmark$ & 3/3 &
0.42 & 0.59 & 25.52 & 14.63 & 46.29 & 11.67  & 1.7 \\
QDrop~\cite{wei2021qdrop} &  $\checkmark$ & 3/3 & 4.44 & 8.00 & 30.73 & 22.67 & 24.37 & 60.89 & 54.76 \\
PD-Quant~\cite{liu2023pd} &  $\checkmark$ & 3/3 & 1.77 & 13.09 & 39.97 & 29.33 & 0.94 & 69.67 & 64.32 \\
RepQ-ViT~\cite{li2023repq} & $\times$ & 3/3 & 0.43  & 0.14 & 0.97 & 4.37 & 4.84 &  8.84 &  1.34\\
I\&S-ViT (Ours) & $\checkmark$ & 3/3 & \textbf{45.16} &  \textbf{63.77} & \textbf{41.52} & \textbf{55.78} & \textbf{73.30} & \textbf{74.20} & \textbf{69.30} \\
\midrule[0.75pt]
FQ-ViT~\cite{lin2022fqvit} &  $\times$ & 4/4 & 0.10 & 0.10 & 0.10 & 0.10 & 0.10 & 0.10  & 0.10 \\
PTQ4ViT~\cite{yuan2022ptq4vit} & $\times$ & 4/4 & 42.57 & 30.69 & 36.96 & 34.08 & 64.39  & 76.09 & 74.02 \\
APQ-ViT~\cite{ding2022towards} &  $\times$ & 4/4 & 47.95 & 41.41 & 47.94 & 43.55 & 67.48 & 77.15 & 76.48 \\
BRECQ~\cite{li2021brecq} &  $\checkmark$ & 4/4 &
12.36 & 9.68 & 55.63 & 63.73 & 72.31  & 72.74  & 58.24 \\
QDrop~\cite{wei2021qdrop} &  $\checkmark$ & 4/4 & 21.24 & 47.30 & 61.93 & 68.27 & 72.60 & 79.58  & 80.93\\
PD-Quant~\cite{liu2023pd} &  $\checkmark$ & 4/4 & 1.51 & 32.45 & 62.46 & 71.21 & 73.76 & 79.87 & 81.12  \\
RepQ-ViT~\cite{li2023repq} & $\times$ & 4/4 & 65.05 & 68.48 &57.43 & 69.03 & 75.61  & 79.45 & 78.32 \\
I\&S-ViT (Ours) & $\checkmark$ & 4/4 & \textbf{74.87}  & \textbf{80.07} & \textbf{65.21}& \textbf{75.81} & \textbf{79.97}  & \textbf{81.17} &  \textbf{82.60}\\
\midrule[0.75pt]
FQ-ViT~\cite{lin2022fqvit} & $\times$ & 6/6 & 4.26 & 0.10 & 58.66 & 45.51 & 64.63  & 66.50 & 52.09 \\
PSAQ-ViT~\cite{li2022patch} & $\times$ & 6/6 & 37.19 & 41.52 & 57.58 & 63.61 & 67.95   & 72.86 & 76.44 \\
Ranking-ViT~\cite{liu2021post} &  $\checkmark$ & 6/6 & - & 75.26 & - & 74.58 & 77.02 &   - & - \\
EasyQuant~\cite{wu2020EasyQuant} &  $\checkmark$ & 6/6 & 75.13 & 81.42 & - & 75.27 & 79.47 &   82.45 & 84.30 \\
PTQ4ViT~\cite{yuan2022ptq4vit} &  $\times$ & 6/6 & 78.63 & 81.65 & 69.68 & 76.28 & 80.25  & 82.38 & 84.01 \\
APQ-ViT \cite{ding2022towards} &  $\times$ & 6/6 & 79.10 & 82.21 & 70.49 & 77.76 & 80.42  & 82.67 & 84.18 \\
NoisyQuant-Linear~\cite{liu2023noisyquant} &  $\times$ & 6/6 & 76.86  & 81.90 & - & 76.37  & 79.77 &    82.78 & 84.57 \\
NoisyQuant-PTQ4ViT~\cite{liu2023noisyquant} &  $\times$ & 6/6 & 78.65  & 82.32 & - & 77.43 & 80.70 &   82.86 & 84.68 \\
BRECQ~\cite{li2021brecq} &  $\checkmark$ & 6/6 & 54.51 & 68.33 & 70.28 & 78.46 & 80.85 & 82.02  & 83.94 \\
QDrop~\cite{wei2021qdrop} &  $\checkmark$ & 6/6 & 70.25 & 75.76 & 70.64 & 77.95 & 80.87 & 82.60 & 84.33\\
PD-Quant~\cite{liu2023pd} &  $\checkmark$ & 6/6 & 70.84 & 75.82 & 70.49 & 78.40 & 80.52 & 82.51 & 84.32\\
Bit-shrinking~\cite{lin2023bit} &  $\checkmark$ & 6/6 & \textbf{80.44} & 83.16 & - & 78.51 & 80.47 & 82.44 & - \\
RepQ-ViT~\cite{li2023repq} & $\times$ & 6/6 & 80.43 & 83.62 & 70.76 & 78.90 & 81.27 &   82.79 &84.57 \\
I\&S-ViT (Ours) & $\checkmark$ & 6/6 & 80.43 & \textbf{83.82} & \textbf{70.85} & \textbf{79.15} & \textbf{81.68}  &   \textbf{82.89} & \textbf{84.94} \\ 
\bottomrule[1.0pt]
\end{tabular}
\caption{Quantization results on ImageNet dataset. The top-1 accuracy (\%) is reported as the metric. ``Opti.'' denotes the optimization-based method, ``Bit. (W/A)'' indicates that the bit-width of the weights and activations are W and A bits, respectively.}
\label{tab:imagenet-0}
\end{table*}

\subsection{Smooth Optimization Strategy}
\label{sec:sos-obs}

It is a wide consensus that post-LayerNorm activations exhibit severe inter-channel variation, necessitating fine-grained quantization granularity~\cite{dong2023packqvit,li2023repq,lin2022fqvit}. \textit{However, the effects of quantization granularity on the optimization process remain underexplored, and in this section, we intend to reveal the internal mechanism}.

In Fig.\,\ref{fig:loss-landscape}, we present the loss landscape when post-LayerNorm activations are subjected to different quantization granularity.
Following ~\cite{frumkin2023jumping}, we plot the loss landscape by adding perturbation to the model weights. Specifically, weights from two random channels are selected, and a basis vector is added to each.
As depicted in Fig.\,\ref{fig:w-q-a-lq}, if the weights undergo channel-wise quantization and post-LayerNorm activations undergo layer-wise quantization, the resulting landscape is rugged and magnified in its loss values. Such an intricate and uneven landscape easily misdirects the learning path into a local minima, which in turn compromises the performance of quantized ViTs~\cite{bai2021binarybert,huang2022sdq,frumkin2023jumping}.
Fortunately, Fig.\,\ref{fig:w-fp-a-lq} suggests that maintaining weights at full-precision results in a significantly smoother loss landscape, albeit a high loss magnitude.
Furthermore, Fig.\,\ref{fig:w-fp-a-cq} showcases that subjecting post-LayerNorm activations to friendly channel-wise quantization ensures not just a gentle and even loss landscape, but one with reduced loss magnitude. 
Such a smooth and low-magnitude loss landscape reduces the learning difficulty~\cite{foret2020sharpness}, establishing a more secure and steadfast foundation upon which the optimization process can well proceed~\cite{li2018visualizing,lin2023bit}.
Spurred by these insights, we introduce a  training strategy, named smooth optimization strategy (SOS), to take advantage of the smooth and low-magnitude loss landscape for optimization at first, while afterward concurrently reaping the benefits of the efficiency proffered by the layer-wise quantizer~\cite{li2023repq,BitSplitStitching,whitepaper}. The proposed SOS comprises three stages, as detailed below:

\textbf{Stage One}. We fine-tune the model while maintaining full-precision weights. At the same time, post-LayerNorm activations are quantized in a channel-wise fashion, according to Fig.\,\ref{fig:w-fp-a-cq}, whereas other activations leverage a layer-wise quantizer. With this setting, the optimization is performed with a smooth loss landscape with lower loss magnitude, thereby establishing a more secure and steadfast learning process.

\textbf{Stage Two}. We employ the scale reparameterization technique~\cite{li2023repq} to realize a transition from the channel-wise quantizer to its layer-wise equivalence. Specifically, 
given the channel-wise scales $\bm{s}\in R^{D}$ and zero-point $\bm{z}\in R^{D}$, $\tilde{s}=\text{Mean}(\bm{s}) \in R^{1}$, $\tilde{z}=\text{Mean}(\bm{z}) \in R^{1}$, $\bm{r}_1=\bm{s}/\tilde{\bm{s}} $, and $\bm{r}_2=\bm{z}-\tilde{\bm{z}}$.
The reparameterization is completed by adjusting the LayerNorm's affine parameters and the weights of the next layer of post-LayerNorm activations:
\begin{equation}
    \label{eq:raparm_1}
  \widetilde{\bm{\beta}} = \frac{\bm{\beta}+\bm{s}\odot \bm{r}_2}{\bm{r}_1}, \quad \widetilde{\bm{\gamma}} = \frac{\bm{\gamma}}{\bm{r}_1}.
\end{equation}
\begin{equation}
\label{eq:raparm_2}
\begin{aligned}
  \widetilde{\bm{W}}_{:,j} = \bm{r}_1\odot\bm{W}_{:,j}, \widetilde{\bm{b}}_j = \bm{b}_j - (\bm{s}\odot \bm{r}_2) \bm{W}_{:,j}.
\end{aligned}
\end{equation}

A detailed analysis can be found in \cite{li2023repq}. Note that, in contrast to prior work that adopts quantized weights $\bm{W}$ and thus introduces lossy transition, our strategy maintains weights at full-precision, ensuring a seamless transition. 

\textbf{Stage Three}. Transitioned weights are quantized and the model undergoes an additional fine-tuning process with quantized activations and weights to restore the performance degradation. 

It is important to note that BRECQ~\cite{li2021brecq} similarly implements a two-stage optimization strategy. In its initial stage, BRECQ conducts optimization using quantized weights alongside full-precision activations, whereas the second stage involves optimization with both being quantized. Nevertheless, our SOS diverges from BRECQ in two fundamental respects:
1) 
Based on the loss landscapes of ViTs, SOS first performs optimization with full-precision weights and quantized activations, while BRECQ is the opposite;
2) SOS incorporates a lossless transition specifically designed to handle high-variant activations special for ViTs, while BRECQ does not consider it.

%
%


%

\begin{table*}[t]
\centering
\small
\begin{tabular}{cccccccccccc}
\toprule[1.25pt]
\multirow{3.5}{*}{\textbf{Method}} &  \multirow{3.5}{*}{\textbf{Opti.}} & \multirow{3.5}{*}{\textbf{Bit. (W/A)}} & \multicolumn{4}{c}{\textbf{Mask R-CNN}} & \multicolumn{4}{c}{\textbf{Cascade Mask R-CNN}} \\
\cmidrule(lr){4-7}\cmidrule(lr){8-11}
&&& \multicolumn{2}{c}{\textbf{w. Swin-T}} & \multicolumn{2}{c}{\textbf{w. Swin-S}} & \multicolumn{2}{c}{\textbf{w. Swin-T}} & \multicolumn{2}{c}{\textbf{w. Swin-S}} \\
&&& AP$^\text{box}$ & AP$^\text{mask}$ & AP$^\text{box}$ & AP$^\text{mask}$ & AP$^\text{box}$ & AP$^\text{mask}$ & AP$^\text{box}$ & AP$^\text{mask}$ \\
\midrule[0.75pt]
Full-Precision &  - & 32/32 & 46.0 & 41.6 & 48.5 & 43.3 & 50.4 & 43.7 & 51.9 & 45.0 \\
\midrule[0.75pt]
PTQ4ViT~\cite{yuan2022ptq4vit} &  $\times$ & 4/4 & 6.9 & 7.0 & 26.7 & 26.6 & 14.7 & 13.5 & 0.5 & 0.5  \\
APQ-ViT~\cite{ding2022towards} &  $\times$ & 4/4 & 23.7 & 22.6 & \textbf{44.7} & 40.1 & 27.2 & 24.4 & 47.7 & 41.1 \\
BRECQ~\cite{li2021brecq} &  $\checkmark$ & 4/4 & 25.4 & 27.6 & 34.9 & 35.4 & 41.2  &  37.0 & 44.5 & 39.2 \\
QDrop~\cite{wei2021qdrop} &  $\checkmark$ & 4/4 & 12.4 & 12.9 & 42.7 & 40.2 & 23.9  & 21.2 & 24.1 & 21.4  \\
PD-Quant~\cite{liu2023pd} &  $\checkmark$ & 4/4 & 17.7 & 18.1 & 32.2  & 30.9 & 35.5  & 31.0 &  41.6 & 36.3 \\
RepQ-ViT~\cite{li2023repq} & $\times$ &  4/4 & 36.1 & 36.0 & 44.2$_{42.7}*$ & 40.2$_{40.1}*$ & 47.0 & 41.4 & 49.3 & 43.1 \\
I\&S-ViT (Ours) &$\checkmark$ & 4/4 &  \textbf{37.5} & \textbf{36.6} & 43.4 &  \textbf{40.3} &  \textbf{48.2} & \textbf{42.0}& \textbf{50.3} & \textbf{43.6}\\
\bottomrule[1.0pt]
\end{tabular}
\caption{Quantization results on COCO dataset. Here, ``AP$^\text{box}$'' denotes the box average precision for object detection, and ``AP$^\text{mask}$'' denotes the mask average precision for instance segmentation. ``*'' indicates the results are reproduced from the official codes.}
\label{tab:coco}
\end{table*}

\section{Experimentation}

\subsection{Experimental Settings}

\paragraph{Models and Datasets}
In order to demonstrate the superiority and generality of I\&S-ViT, we subject it to rigorous evaluation across diverse visual tasks, including image classification, object detection, and instance segmentation.
For the image classification task, we evaluate the I\&S-ViT on the ImageNet dataset~\cite{russakovsky2015imagenet}, considering different model variants including ViT~\cite{DosovitskiyZ21An}, DeiT~\cite{touvron2021training}, and Swin~\cite{liu2021swin}.
For object detection and instance segmentation tasks, we evaluate I\&S-ViT on the COCO dataset \cite{lin2014microsoft} using two prevalent frameworks: Mask R-CNN~\cite{he2017mask} and Cascade Mask R-CNN \cite{cai2018cascade}, both with Swin~\cite{liu2021swin} as the backbone.

\paragraph{Implementation details} 
All experiments are executed utilizing the PyTorch framework~\cite{paszke2019pytorch}, with pre-trained full-precision models sourced from the Timm library.
We adopt the uniform quantizer for all weights and activations except for the post-Softmax activations, which are handled by the proposed shift-uniform-log2 quantizer. We adopt the straight-through estimator(STE)~\cite{courbariaux2016binarized} to bypass the calculation of the gradient of the non-differentiable rounding function.
Consistent with preceding studies~\cite{frumkin2023jumping,liu2023noisyquant}, we arbitrarily select 1024 images each from the ImageNet and COCO datasets. The Adam optimizer~\cite{kingma2014adam} is employed for optimization. The initialized learning rate is 4e-5 for weights, with weight decay set to 0. The learning rate undergoes adjustment via the cosine learning rate decay strategy. As pointed out in \cite{frumkin2023jumping,ding2022towards}, the quantization parameters yield numerous local minima in the loss landscape, easily misleading the learning direction. Thus, we do not optimize them after calibration.
For the ImageNet dataset, the batch size is 64 and the training iteration is 200  for the 6-bit case and 1000 for other cases. 
%
For the COCO dataset, we only optimize the backbone, and the remaining structures are quantized with the calibration strategy as in \cite{li2023repq}. A batch size of 1 with a training iteration of 1000 is used. 
In our experiments, SULQ' $\eta$ is determined before the optimization process by grid searching the one with the minimum quantization error from candidates. 
All experiments are implemented using a single NVIDIA 3090 GPU.

\subsection{Results on ImageNet Dataset}

The comparison between the proposed I\&S-ViT and other PTQ of ViTs methods is reported in Tab.\,\ref{tab:imagenet-0}.

Specifically, the advantages of our I\&S-ViT are highlighted in all bit cases, especially for low-bit cases. As illustrated in Tab.\,\ref{tab:imagenet-0}, both optimization-free and optimization-based methods suffer from non-trivial performance degradation in the ultra-low bit cases.
For instance, in the 3-bit case, optimization-based PTQ4ViT~\cite{yuan2022ptq4vit} suffers from collapse for all ViT variants, and RepQ-ViT presents limited accuracy.
For instance, RepQ-ViT only presents 0.97\%, 4.37\%, and 4.84\% for DeiT-T, DeiT-B, and DeiT-B, respectively.
The optimization-based methods present better results but showcase an unstable performance for different ViT variants. For example, BRECQ~\cite{li2021brecq} suffers from collapse on ViT-S and Swin-B.
In contrast, the proposed I\&S-ViT showcases a stable and considerably improved performance for ViT variants.
In particular, I\&S-ViT respectively presents an encouraging 40.72\% and 50.68\% improvement over previous methods in ViT-S and ViT-B quantization.
On DeiT-T, DeiT-B, and DeiT-B, I\&S-ViT respectively obtain 41.52\%, 55.78\%, and 73.30\% performance, respectively corresponding to 1.55\%, 26.45\%, and 27.01\% increases.
On Swin-S and Swin-B, I\&S-ViT reports 4.53\% and 4.98\% increases, respectively.

In the 4-bit case, the optimization-free RepQ-ViT outperforms optimization-based methods on most ViT variants, demonstrating that previous optimization-based PTQ methods suffer from the overfitting issue.
While the proposed I\&S-ViT presents considerable improvements over RepQ-ViT across ViT variants. Specifically, I\&S-ViT achieves notable 9.82\% and 11.59\% improvements for ViT-S and ViT-B, respectively. 
When quantizing DeiT-T, DeiT-S, and DeiT-B, I\&S-ViT provides notable 3.28\%, 6.78\%, and 4.36\% accuracy gains, respectively.
As for Swin-S and Swin-B, I\&S-ViT showcases 1.72\% and 1.48\% performance gains, respectively.

In the 6-bit case, RepQ-ViT outperforms optimization-based methods for most ViT variants, indicating that optimization-based methods also suffer from the same overfitting issue as in the 4-bit case.
Similar to the results on the 3-bit and 4-bit cases, I\&S-ViT presents performance improvements and satisfactory results. For instance, in DeiT-B, Swin-S, and Swin-B quantization, I\&S-ViT presents 81.68\%, 82.89\%, and 84.94\% accuracy, respectively, with only 0.12\%, 0.34\%, and 0.33\% accuracy loss compared with the full-precision model.

\begin{table}[ht]
\centering
\begin{tabular}{ccccc}
\toprule[1.25pt]
\textbf{Model}                                                               & \textbf{SULQ} & \textbf{SOS }& \textbf{Top-1 Acc. (\%)} \\ \midrule[0.75pt]
\multirow{5}{*}{\begin{tabular}[c]{@{}c@{}}DeiT-S (W3A3)  \end{tabular}}   & \multicolumn{2}{c}{Full-Precision}     &   79.85  \\ \cline{2-4}   
                                                                                  &             &   & 3.36 \\
                                                                &     $\checkmark$   &             &  20.70               \\
                                                                                    &           &  $\checkmark$  &  45.19    \\
                                                                                    &   $\checkmark$        &  $\checkmark$  &  \textbf{55.78}   \\
\bottomrule[1.0pt]                                                                    
\end{tabular}
\caption{Ablation studies of the effectiveness of shift-uniform-log2 quantizer (SULQ) and the smooth optimization strategy (SOS).}
\label{tab:ablation-components}
\end{table}

\begin{table}[t]
\centering
\begin{tabular}{ccc}
\toprule[1.25pt]
\textbf{Model} & \textbf{Method}  &\textbf{Top-1 Acc. (\%)} \\
\midrule[0.75pt]
\multirow{4.5}{*}{DeiT-S (W3/A3)} & Full-Precision & 79.85 \\
\cmidrule{2-3}
& LQ  &  52.60 \\
& UQ  &  44.79 \\
& SULQ (Ours)  &  \textbf{55.78} \\
\bottomrule[1.0pt]
\end{tabular}
\caption{Ablation studies of different quantizers for post-Softmax activations. ``LQ'' and ``UQ'' denote the log2 quantizer and the uniform quantizer, respectively.}
\label{tab:ablation-postSoft}
\end{table}

\subsection{Results on COCO Dataset}

The results of object detection and instance segmentation are reported in Tab.\,\ref{tab:coco}.
All networks are quantized to 4-bit.
It can be seen that I\&S-ViT achieves a better performance in most cases. 
To be specific, when Mask R-CNN employs Swin-T as its backbone, I\&S-ViT augments the box AP and mask AP by 1.4 and 0.6 points, respectively. Similarly, with Cascade Mask R-CNN, I\&S-ViT enhances the box AP by 1.2 and mask AP by 0.6 when Swin-T serves as the backbone. When Swin-S is utilized as the backbone, the improvements are 1.0 for box AP and 0.5 for mask AP.

\begin{figure}[!ht]
\centering
\includegraphics[width=0.85\linewidth]{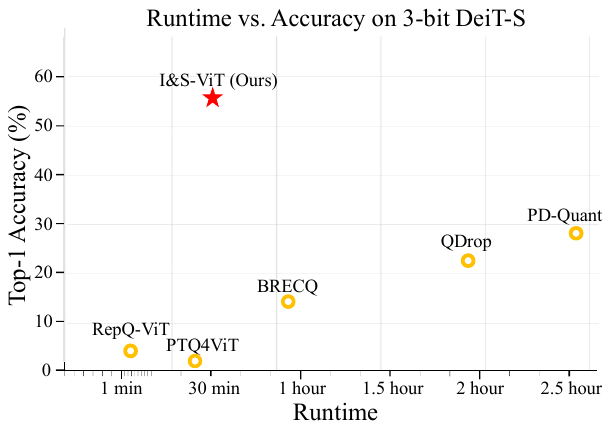} 
\caption{The accuracy vs. runtime of PTQ methods on 3-bit DeiT.}
\label{fig:timecosts}
\end{figure}

\subsection{Ablation Studies}
\label{sec:ablation}

\textbf{Effect of SULQ and SOS} 
Tab.\,\ref{tab:ablation-components} reports the ablation study of the proposed shift-uniform-log2 quantizer (SULQ) and the smooth optimization strategy (SOS). If SULQ is not used, we utilize the log2 quantizer as an alternative. 
As can be observed, the proposed SULQ and SOS both contribute to the performance considerably. If both SULQ and SOS are removed, 3-bit DeiT-S only yields 3.36\%. Applying SULQ improved the accuracy by 17.34\%. By using SOS, 3-bit DeiT-S yields 45.19\% accuracy. At last, when both SULQ and SOS are adopted, it presents the best performance, \emph{i.e.}, 55.78\% for 3-bit DeiT-S.

\textbf{Effect of SULQ for post-Softmax activations} 
Tab.\,\ref{tab:ablation-postSoft} reports the accuracy of different quantizers for post-Softmax activations.
As can be seen, if using the uniform quantizer, 3-bit DeiT-S suffers from 3.18\% accuracy degradation. When using the log2 quantizer, 3-bit DeiT-S suffers from 10.99\% accuracy drops. In contrast, the proposed SULQ presents an improved performance, demonstrating its superiority.

\textbf{Time efficiency}
Fig.\,\ref{fig:timecosts} showcases the runtime comparison. Notably, the proposed I\&S-ViT significantly outperforms all other PTQ4 methods while maintaining a decent time cost. I\&S-ViT roughly consumes 31 minutes. Compared with optimization-based BRECQ, QDdrop, and PD-Quant, the time cost of I\&S-ViT is only about one-half to one-fifth of the consumption.
Compared with optimization-free RepQ-ViT and PTQ4ViT, the consumed time of I\&S-ViT remains in the same magnitude.

\section{Discussion}

While the proposed I\&S-ViT substantially enhances the performance of PTQ for ViTs, a gap persists between the quantized model and its full-precision counterpart in the low-bit scenarios. It remains crucial to identify a more effective PTQ method tailored for ViTs. For instance, block-wise optimization might not be the optimal solution; thus, exploring finer-grained granularity for optimization targets could be beneficial. Moreover, even though the SULQ designed for post-Softmax activations demonstrates commendable performance and adaptability, the quest for an even more efficient quantizer remains a valuable avenue of exploration. We hope the proposed I\&S-ViT could serve as a strong baseline for future researchers in this domain.

\section{Conclusion}

In this paper, we introduced I\&S-ViT, a novel optimized-based PTQ method tailored specifically for ViTs. 
At the outset, we address the quantization inefficiency issue associated with the log2 quantizer by introducing the shift-uniform-log2 quantizer (SULQ).
The SULQ inclusively represents the full input domain to effectively address the quantization inefficiency issue and accurately approximate the distributions of post-Softmax activations.
Then, our insights into the contrasting loss landscapes of different quantization granularity, guide the development of the three-stage smooth optimization strategy (SOS). SOS enables stable learning by exploiting the smooth and low-magnitude loss landscape of channel-wise quantization for optimization while presenting efficiency by utilizing layer-wise quantization through seamless scale reparameterization.
The superiority of I\&S-ViT is demonstrated by extensive experiments on various ViTs of different vision tasks.

\noindent\textbf{Acknowledgements}. This work was supported by National Key R\&D Program of China (No.2022ZD0118202), the National Science Fund for Distinguished Young Scholars (No.62025603), the National Natural Science Foundation of China (No. U21B2037, No. U22B2051, No. 62176222, No. 62176223, No. 62176226, No. 62072386, No. 62072387, No. 62072389, No. 62002305 and No. 62272401), and the Natural Science Foundation of Fujian Province of China (No.2021J01002,  No.2022J06001).

{
    \small
    \bibliographystyle{ieeenat_fullname}
    \bibliography{main}
}


\clearpage
\appendix

\section*{Appendix \label{appendix}}

\section{Ablation Studies}

\begin{figure}[!ht]
\centering
\includegraphics[width=0.85\linewidth]{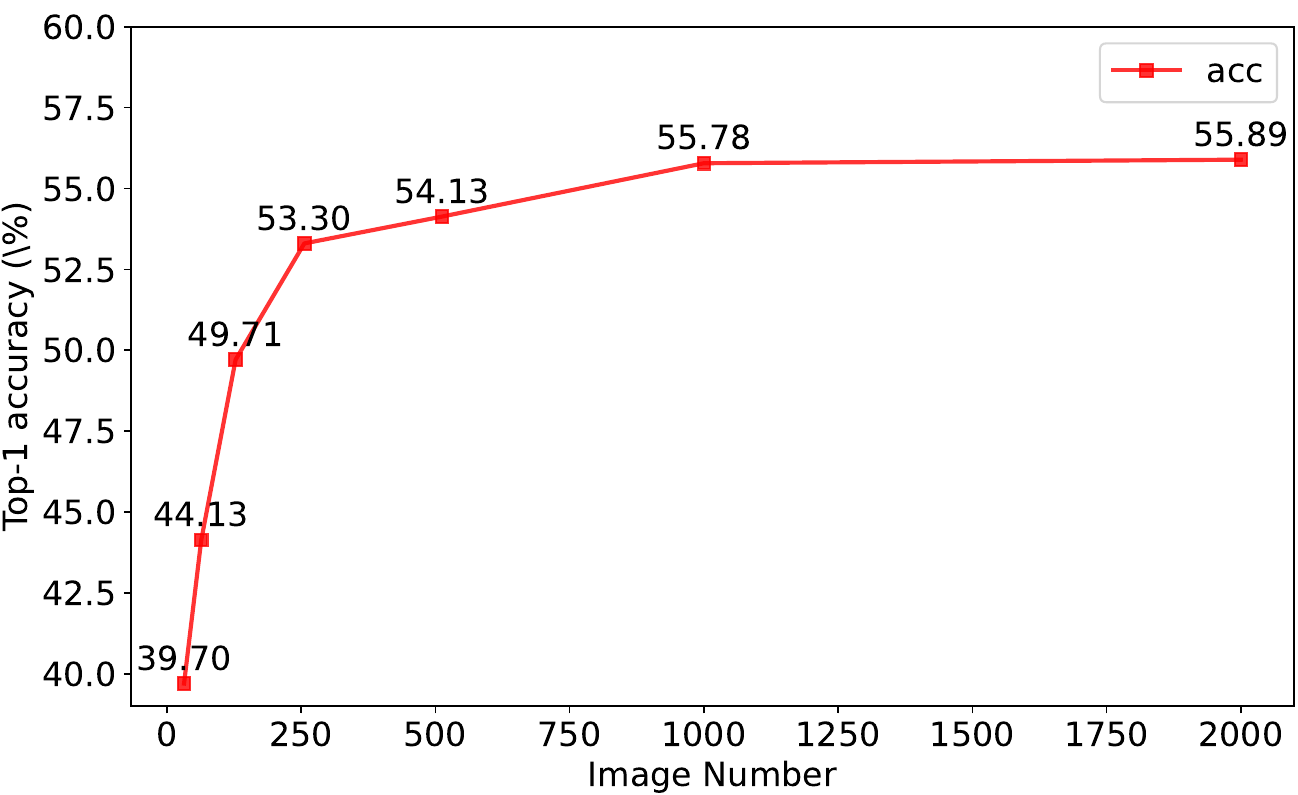} 
\caption{The accuracy vs. image number on 3-bit DeiT.}
\label{fig:imagenumber}
\end{figure}

\textbf{Effect of image number} 
Fig\,\ref{fig:imagenumber} reports the ablation study of different image numbers. 
As can be observed, when using 32 images, the top-1 accuracy is 39.70\%. As the number increases, the performance is improved. For 512 images, the performance is 54.13\%. When it comes to 1024 images, which also is the setting in our main paper, the top-1 accuracy is 55.78\%. Afterward, continually using more images does not bring a significant performance boost as it presents only 55.89\% for 2048 images.

\end{document}

%% file: preamble.tex
\usepackage[dvipsnames]{xcolor}


%% file: main.bbl
\begin{thebibliography}{61}
\providecommand{\natexlab}[1]{#1}
\providecommand{\url}[1]{\texttt{#1}}
\expandafter\ifx\csname urlstyle\endcsname\relax
  \providecommand{\doi}[1]{doi: #1}\else
  \providecommand{\doi}{doi: \begingroup \urlstyle{rm}\Url}\fi

\bibitem[Arnab et~al.(2021)Arnab, Dehghani, Heigold, Sun, Lu{\v{c}}i{\'c}, and Schmid]{arnab2021vivit}
Anurag Arnab, Mostafa Dehghani, Georg Heigold, Chen Sun, Mario Lu{\v{c}}i{\'c}, and Cordelia Schmid.
\newblock Vivit: A video vision transformer.
\newblock In \emph{Proceedings of the IEEE/CVF international conference on computer vision (ICCV)}, pages 6836--6846, 2021.

\bibitem[Bai et~al.(2021)Bai, Zhang, Hou, Shang, Jin, Jiang, Liu, Lyu, and King]{bai2021binarybert}
Haoli Bai, Wei Zhang, Lu Hou, Lifeng Shang, Jin Jin, Xin Jiang, Qun Liu, Michael~R. Lyu, and Irwin King.
\newblock Binarybert: Pushing the limit of {BERT} quantization.
\newblock In \emph{Proceedings of the 59th Annual Meeting of the Association for Computational Linguistics and the 11th International Joint Conference on Natural Language Processing, {ACL/IJCNLP} 2021, (Volume 1: Long Papers), Virtual Event, August 1-6, 2021}, pages 4334--4348, 2021.

\bibitem[Banner et~al.(2019)Banner, Nahshan, Soudry, et~al.]{ACIQ}
Ron Banner, Yury Nahshan, Daniel Soudry, et~al.
\newblock Post training 4-bit quantization of convolutional networks for rapid-deployment.
\newblock In \emph{Proceedings of the Advances in Neural Information Processing Systems (NeurIPS)}, pages 7950--7958, 2019.

\bibitem[Bolya et~al.(2023)Bolya, Fu, Dai, Zhang, Feichtenhofer, and Hoffman]{bolya2022token}
Daniel Bolya, Cheng-Yang Fu, Xiaoliang Dai, Peizhao Zhang, Christoph Feichtenhofer, and Judy Hoffman.
\newblock Token merging: Your vit but faster.
\newblock In \emph{The Eleventh International Conference on Learning Representations (ICLR)}, 2023.

\bibitem[Cai et~al.(2018)Cai, Takemoto, and Nakajo]{cai2018deep}
Jingyong Cai, Masashi Takemoto, and Hironori Nakajo.
\newblock A deep look into logarithmic quantization of model parameters in neural networks.
\newblock In \emph{Proceedings of the Advances in Neural Information Processing Systems (NeurIPS)}, pages 1--8, 2018.

\bibitem[Cai and Vasconcelos(2018)]{cai2018cascade}
Zhaowei Cai and Nuno Vasconcelos.
\newblock Cascade r-cnn: Delving into high quality object detection.
\newblock In \emph{Proceedings of the IEEE/CVF Conference on Computer Vision and Pattern Recognition (CVPR)}, pages 6154--6162, 2018.

\bibitem[Carion et~al.(2020)Carion, Massa, Synnaeve, Usunier, Kirillov, and Zagoruyko]{carion2020end}
Nicolas Carion, Francisco Massa, Gabriel Synnaeve, Nicolas Usunier, Alexander Kirillov, and Sergey Zagoruyko.
\newblock End-to-end object detection with transformers.
\newblock In \emph{Proceedings of the European Conference on Computer Vision (ECCV)}, pages 213--229. Springer, 2020.

\bibitem[Chen et~al.(2023)Chen, Shao, Xu, Lin, Zhang, Chao, Ji, Qiao, and Luo]{chen2023diffrate}
Mengzhao Chen, Wenqi Shao, Peng Xu, Mingbao Lin, Kaipeng Zhang, Fei Chao, Rongrong Ji, Yu Qiao, and Ping Luo.
\newblock Diffrate: Differentiable compression rate for efficient vision transformers.
\newblock \emph{arXiv preprint arXiv:2305.17997}, 2023.

\bibitem[Courbariaux et~al.(2016)Courbariaux, Hubara, Soudry, El-Yaniv, and Bengio]{courbariaux2016binarized}
Matthieu Courbariaux, Itay Hubara, Daniel Soudry, Ran El-Yaniv, and Yoshua Bengio.
\newblock Binarized neural networks: Training deep neural networks with weights and activations constrained to+ 1 or-1.
\newblock \emph{arXiv preprint arXiv:1602.02830}, 2016.

\bibitem[Ding et~al.(2022)Ding, Qin, Yan, Chai, Liu, Wei, and Liu]{ding2022towards}
Yifu Ding, Haotong Qin, Qinghua Yan, Zhenhua Chai, Junjie Liu, Xiaolin Wei, and Xianglong Liu.
\newblock Towards accurate post-training quantization for vision transformer.
\newblock In \emph{Proceedings of the 30th ACM International Conference on Multimedia (ACMMM)}, pages 5380--5388, 2022.

\bibitem[Dong et~al.(2023)Dong, Lu, Wu, Lyu, Yuan, Tang, and Wang]{dong2023packqvit}
Peiyan Dong, Lei Lu, Chao Wu, Cheng Lyu, Geng Yuan, Hao Tang, and Yanzhi Wang.
\newblock Packqvit: Faster sub-8-bit vision transformers via full and packed quantization on the mobile.
\newblock In \emph{Proceedings of the Advances in Neural Information Processing Systems (NeurIPS)}, 2023.

\bibitem[Dosovitskiy et~al.(2021)Dosovitskiy, Beyer, Kolesnikov, Weissenborn, Zhai, Unterthiner, Dehghani, Minderer, Heigold, Gelly, Uszkoreit, and Houlsby]{DosovitskiyZ21An}
Alexey Dosovitskiy, Lucas Beyer, Alexander Kolesnikov, Dirk Weissenborn, Xiaohua Zhai, Thomas Unterthiner, Mostafa Dehghani, Matthias Minderer, Georg Heigold, Sylvain Gelly, Jakob Uszkoreit, and Neil Houlsby.
\newblock An image is worth 16x16 words: Transformers for image recognition at scale.
\newblock In \emph{Proceedings of the International Conference on Learning Representations (ICLR)}. OpenReview.net, 2021.

\bibitem[Esser et~al.(2020)Esser, McKinstry, Bablani, Appuswamy, and Modha]{LSQ}
Steven~K. Esser, Jeffrey~L. McKinstry, Deepika Bablani, Rathinakumar Appuswamy, and Dharmendra~S. Modha.
\newblock Learned step size quantization.
\newblock In \emph{Proceedings of the International Conference on Learning Representations (ICLR)}, 2020.

\bibitem[Foret et~al.(2021)Foret, Kleiner, Mobahi, and Neyshabur]{foret2020sharpness}
Pierre Foret, Ariel Kleiner, Hossein Mobahi, and Behnam Neyshabur.
\newblock Sharpness-aware minimization for efficiently improving generalization.
\newblock In \emph{The Eleventh International Conference on Learning Representations (ICLR)}, 2021.

\bibitem[Frumkin et~al.(2023)Frumkin, Gope, and Marculescu]{frumkin2023jumping}
Natalia Frumkin, Dibakar Gope, and Diana Marculescu.
\newblock Jumping through local minima: Quantization in the loss landscape of vision transformers.
\newblock In \emph{Proceedings of the IEEE/CVF International Conference on Computer Vision (ICCV)}, pages 16978--16988, 2023.

\bibitem[Gong et~al.(2019)Gong, Liu, Jiang, Li, Hu, Lin, Yu, and Yan]{gong2019differentiable}
Ruihao Gong, Xianglong Liu, Shenghu Jiang, Tianxiang Li, Peng Hu, Jiazhen Lin, Fengwei Yu, and Junjie Yan.
\newblock Differentiable soft quantization: Bridging full-precision and low-bit neural networks.
\newblock In \emph{Proceedings of the IEEE/CVF Conference on Computer Vision and Pattern Recognition (CVPR)}, pages 4852--4861, 2019.

\bibitem[He et~al.(2017)He, Gkioxari, Doll{\'a}r, and Girshick]{he2017mask}
Kaiming He, Georgia Gkioxari, Piotr Doll{\'a}r, and Ross Girshick.
\newblock Mask r-cnn.
\newblock In \emph{Proceedings of the IEEE/CVF International Conference on Computer Vision (ICCV)}, pages 2961--2969, 2017.

\bibitem[Huang et~al.(2022)Huang, Shen, Li, Liu, Xianghong, Wicaksana, Xing, and Cheng]{huang2022sdq}
Xijie Huang, Zhiqiang Shen, Shichao Li, Zechun Liu, Hu Xianghong, Jeffry Wicaksana, Eric Xing, and Kwang-Ting Cheng.
\newblock Sdq: Stochastic differentiable quantization with mixed precision.
\newblock In \emph{Proceedings of the International Conference on Machine Learning (ICML)}, pages 9295--9309. PMLR, 2022.

\bibitem[Jung et~al.(2019)Jung, Son, Lee, Son, Han, Kwak, Hwang, and Choi]{jung2019learning}
Sangil Jung, Changyong Son, Seohyung Lee, Jinwoo Son, Jae-Joon Han, Youngjun Kwak, Sung~Ju Hwang, and Changkyu Choi.
\newblock Learning to quantize deep networks by optimizing quantization intervals with task loss.
\newblock In \emph{Proceedings of the IEEE/CVF Conference on Computer Vision and Pattern Recognition (CVPR)}, pages 4350--4359, 2019.

\bibitem[Kingma and Ba(2014)]{kingma2014adam}
Diederik~P Kingma and Jimmy Ba.
\newblock Adam: A method for stochastic optimization.
\newblock In \emph{Proceedings of the International Conference on Learning Representations (ICLR)}, 2014.

\bibitem[Krishnamoorthi(2018)]{whitepaper}
Raghuraman Krishnamoorthi.
\newblock Quantizing deep convolutional networks for efficient inference: A whitepaper.
\newblock \emph{arXiv preprint arXiv:1806.08342}, 2018.

\bibitem[Li et~al.(2018)Li, Xu, Taylor, Studer, and Goldstein]{li2018visualizing}
Hao Li, Zheng Xu, Gavin Taylor, Christoph Studer, and Tom Goldstein.
\newblock Visualizing the loss landscape of neural nets.
\newblock In \emph{Proceedings of the Advances in Neural Information Processing Systems (NeurIPS)}, 2018.

\bibitem[Li et~al.(2020)Li, Dong, and Wang]{APoT}
Yuhang Li, Xin Dong, and Wei Wang.
\newblock Additive powers-of-two quantization: An efficient non-uniform discretization for neural networks.
\newblock In \emph{Proceedings of the International Conference on Learning Representations (ICLR)}, 2020.

\bibitem[Li et~al.(2021)Li, Gong, Tan, Yang, Hu, Zhang, Yu, Wang, and Gu]{li2021brecq}
Yuhang Li, Ruihao Gong, Xu Tan, Yang Yang, Peng Hu, Qi Zhang, Fengwei Yu, Wei Wang, and Shi Gu.
\newblock Brecq: Pushing the limit of post-training quantization by block reconstruction.
\newblock In \emph{Proceedings of the International Conference on Learning Representations (ICLR)}, 2021.

\bibitem[Li et~al.(2022{\natexlab{a}})Li, Xu, Zhang, Cao, Gao, and Guo]{li2022q}
Yanjing Li, Sheng Xu, Baochang Zhang, Xianbin Cao, Peng Gao, and Guodong Guo.
\newblock Q-vit: Accurate and fully quantized low-bit vision transformer.
\newblock In \emph{Proceedings of the Advances in Neural Information Processing Systems (NeurIPS)}, pages 34451--34463, 2022{\natexlab{a}}.

\bibitem[Li and Gu(2023)]{li2023vit}
Zhikai Li and Qingyi Gu.
\newblock I-vit: Integer-only quantization for efficient vision transformer inference.
\newblock In \emph{Proceedings of the IEEE/CVF International Conference on Computer Vision (ICCV)}, pages 17065--17075, 2023.

\bibitem[Li et~al.(2022{\natexlab{b}})Li, Ma, Chen, Xiao, and Gu]{li2022patch}
Zhikai Li, Liping Ma, Mengjuan Chen, Junrui Xiao, and Qingyi Gu.
\newblock Patch similarity aware data-free quantization for vision transformers.
\newblock In \emph{Proceedings of the European Conference on Computer Vision (ECCV)}, pages 154--170. Springer, 2022{\natexlab{b}}.

\bibitem[Li et~al.(2022{\natexlab{c}})Li, Yang, Wang, and Cheng]{li2022qv}
Zhexin Li, Tong Yang, Peisong Wang, and Jian Cheng.
\newblock Q-vit: Fully differentiable quantization for vision transformer.
\newblock \emph{CoRR}, abs/2201.07703, 2022{\natexlab{c}}.

\bibitem[Li et~al.(2023)Li, Xiao, Yang, and Gu]{li2023repq}
Zhikai Li, Junrui Xiao, Lianwei Yang, and Qingyi Gu.
\newblock Repq-vit: Scale reparameterization for post-training quantization of vision transformers.
\newblock In \emph{Proceedings of the IEEE/CVF International Conference on Computer Vision (ICCV)}, pages 17227--17236, 2023.

\bibitem[Liang et~al.(2021)Liang, Cao, Sun, Zhang, Van~Gool, and Timofte]{liang2021swinir}
Jingyun Liang, Jiezhang Cao, Guolei Sun, Kai Zhang, Luc Van~Gool, and Radu Timofte.
\newblock Swinir: Image restoration using swin transformer.
\newblock In \emph{Proceedings of the IEEE/CVF international conference on computer vision (ICCV)}, pages 1833--1844, 2021.

\bibitem[Lin et~al.(2023{\natexlab{a}})Lin, Peng, Li, Tan, Ren, Xiao, and Pu]{lin2023bit}
Chen Lin, Bo Peng, Zheyang Li, Wenming Tan, Ye Ren, Jun Xiao, and Shiliang Pu.
\newblock Bit-shrinking: Limiting instantaneous sharpness for improving post-training quantization.
\newblock In \emph{Proceedings of the IEEE/CVF Conference on Computer Vision and Pattern Recognition (CVPR)}, pages 16196--16205, 2023{\natexlab{a}}.

\bibitem[Lin et~al.(2023{\natexlab{b}})Lin, Chen, Zhang, Shen, Ji, and Cao]{lin2023super}
Mingbao Lin, Mengzhao Chen, Yuxin Zhang, Chunhua Shen, Rongrong Ji, and Liujuan Cao.
\newblock Super vision transformer.
\newblock \emph{International Journal of Computer Vision (IJCV)}, pages 1--16, 2023{\natexlab{b}}.

\bibitem[Lin et~al.(2014)Lin, Maire, Belongie, Hays, Perona, Ramanan, Doll{\'a}r, and Zitnick]{lin2014microsoft}
Tsung-Yi Lin, Michael Maire, Serge Belongie, James Hays, Pietro Perona, Deva Ramanan, Piotr Doll{\'a}r, and C~Lawrence Zitnick.
\newblock Microsoft coco: Common objects in context.
\newblock In \emph{Proceedings of the European Conference on Computer Vision (ECCV)}, pages 740--755. Springer, 2014.

\bibitem[Lin et~al.(2022)Lin, Zhang, Sun, Li, and Zhou]{lin2022fqvit}
Yang Lin, Tianyu Zhang, Peiqin Sun, Zheng Li, and Shuchang Zhou.
\newblock Fq-vit: Post-training quantization for fully quantized vision transformer.
\newblock In \emph{Proceedings of the Thirty-First International Joint Conference on Artificial Intelligence, (IJCAI)}, pages 1173--1179, 2022.

\bibitem[Liu et~al.(2023{\natexlab{a}})Liu, Niu, Yuan, Yang, Wang, and Liu]{liu2023pd}
Jiawei Liu, Lin Niu, Zhihang Yuan, Dawei Yang, Xinggang Wang, and Wenyu Liu.
\newblock Pd-quant: Post-training quantization based on prediction difference metric.
\newblock In \emph{Proceedings of the IEEE/CVF Conference on Computer Vision and Pattern Recognition (CVPR)}, pages 24427--24437, 2023{\natexlab{a}}.

\bibitem[Liu et~al.(2023{\natexlab{b}})Liu, Liu, and Cheng]{Yang2023Oscillation}
Shih{-}Yang Liu, Zechun Liu, and Kwang{-}Ting Cheng.
\newblock Oscillation-free quantization for low-bit vision transformers.
\newblock In \emph{Proceedings of the International Conference on Machine Learning (ICML)}, pages 21813--21824, 2023{\natexlab{b}}.

\bibitem[Liu et~al.(2023{\natexlab{c}})Liu, Yang, Dong, Keutzer, Du, and Zhang]{liu2023noisyquant}
Yijiang Liu, Huanrui Yang, Zhen Dong, Kurt Keutzer, Li Du, and Shanghang Zhang.
\newblock Noisyquant: Noisy bias-enhanced post-training activation quantization for vision transformers.
\newblock In \emph{Proceedings of the IEEE/CVF Conference on Computer Vision and Pattern Recognition (CVPR)}, pages 20321--20330, 2023{\natexlab{c}}.

\bibitem[Liu et~al.(2021{\natexlab{a}})Liu, Lin, Cao, Hu, Wei, Zhang, Lin, and Guo]{liu2021swin}
Ze Liu, Yutong Lin, Yue Cao, Han Hu, Yixuan Wei, Zheng Zhang, Stephen Lin, and Baining Guo.
\newblock Swin transformer: Hierarchical vision transformer using shifted windows.
\newblock In \emph{Proceedings of the IEEE/CVF international conference on computer vision (ICCV)}, pages 10012--10022, 2021{\natexlab{a}}.

\bibitem[Liu et~al.(2021{\natexlab{b}})Liu, Wang, Han, Zhang, Ma, and Gao]{liu2021post}
Zhenhua Liu, Yunhe Wang, Kai Han, Wei Zhang, Siwei Ma, and Wen Gao.
\newblock Post-training quantization for vision transformer.
\newblock In \emph{Proceedings of the Advances in Neural Information Processing Systems (NeurIPS)}, pages 28092--28103, 2021{\natexlab{b}}.

\bibitem[Mehta and Rastegari(2022)]{mehta2021mobilevit}
Sachin Mehta and Mohammad Rastegari.
\newblock Mobilevit: Light-weight, general-purpose, and mobile-friendly vision transformer.
\newblock In \emph{Proceedings of the International Conference on Learning Representations (ICLR)}, 2022.

\bibitem[Nagel et~al.(2020)Nagel, Amjad, Van~Baalen, Louizos, and Blankevoort]{Upordown}
Markus Nagel, Rana~Ali Amjad, Mart Van~Baalen, Christos Louizos, and Tijmen Blankevoort.
\newblock Up or down? adaptive rounding for post-training quantization.
\newblock In \emph{Proceedings of the International Conference on Machine Learning (ICML)}, pages 7197--7206, 2020.

\bibitem[Paszke et~al.(2019)Paszke, Gross, Massa, Lerer, Bradbury, Chanan, Killeen, Lin, Gimelshein, Antiga, et~al.]{paszke2019pytorch}
Adam Paszke, Sam Gross, Francisco Massa, Adam Lerer, James Bradbury, Gregory Chanan, Trevor Killeen, Zeming Lin, Natalia Gimelshein, Luca Antiga, et~al.
\newblock Pytorch: An imperative style, high-performance deep learning library.
\newblock In \emph{Proceedings of the Advances in Neural Information Processing Systems (NeurIPS)}, pages 8026--8037, 2019.

\bibitem[Qin et~al.(2020)Qin, Gong, Liu, Shen, Wei, Yu, and Song]{qin2020forward}
Haotong Qin, Ruihao Gong, Xianglong Liu, Mingzhu Shen, Ziran Wei, Fengwei Yu, and Jingkuan Song.
\newblock Forward and backward information retention for accurate binary neural networks.
\newblock In \emph{Proceedings of the IEEE/CVF Conference on Computer Vision and Pattern Recognition (CVPR)}, pages 2250--2259, 2020.

\bibitem[Russakovsky et~al.(2015)Russakovsky, Deng, Su, Krause, Satheesh, Ma, Huang, Karpathy, Khosla, Bernstein, et~al.]{russakovsky2015imagenet}
Olga Russakovsky, Jia Deng, Hao Su, Jonathan Krause, Sanjeev Satheesh, Sean Ma, Zhiheng Huang, Andrej Karpathy, Aditya Khosla, Michael Bernstein, et~al.
\newblock Imagenet large scale visual recognition challenge.
\newblock \emph{International Journal of Computer Vision (IJCV)}, 115:\penalty0 211--252, 2015.

\bibitem[Touvron et~al.(2021)Touvron, Cord, Douze, Massa, Sablayrolles, and J{\'e}gou]{touvron2021training}
Hugo Touvron, Matthieu Cord, Matthijs Douze, Francisco Massa, Alexandre Sablayrolles, and Herv{\'e} J{\'e}gou.
\newblock Training data-efficient image transformers \& distillation through attention.
\newblock In \emph{Proceedings of the International Conference on Machine Learning (ICML)}, pages 10347--10357. PMLR, 2021.

\bibitem[Vaswani et~al.(2017)Vaswani, Shazeer, Parmar, Uszkoreit, Jones, Gomez, Kaiser, and Polosukhin]{vaswani2017attention}
Ashish Vaswani, Noam Shazeer, Niki Parmar, Jakob Uszkoreit, Llion Jones, Aidan~N Gomez, {\L}ukasz Kaiser, and Illia Polosukhin.
\newblock Attention is all you need.
\newblock In \emph{Proceedings of the International Conference on Neural Information Processing Systems (NeurIPS)}, pages 6000--6010, 2017.

\bibitem[Wang et~al.(2020)Wang, Chen, He, and Cheng]{BitSplitStitching}
Peisong Wang, Qiang Chen, Xiangyu He, and Jian Cheng.
\newblock Towards accurate post-training network quantization via bit-split and stitching.
\newblock In \emph{Proceedings of the International Conference on Machine Learning (ICML)}, pages 9847--9856, 2020.

\bibitem[Wei et~al.(2022)Wei, Gong, Li, Liu, and Yu]{wei2021qdrop}
Xiuying Wei, Ruihao Gong, Yuhang Li, Xianglong Liu, and Fengwei Yu.
\newblock Qdrop: Randomly dropping quantization for extremely low-bit post-training quantization.
\newblock In \emph{Proceedings of the International Conference on Learning Representations (ICLR)}, 2022.

\bibitem[Wu et~al.(2020)Wu, Tang, Zhao, Zhang, Fu, and Zhang]{wu2020EasyQuant}
Di Wu, Qi Tang, Yongle Zhao, Ming Zhang, Ying Fu, and Debing Zhang.
\newblock Easyquant: Post-training quantization via scale optimization.
\newblock \emph{CoRR}, abs/2006.16669, 2020.

\bibitem[Xu et~al.(2021)Xu, Lin, Liu, Chen, Shao, Gao, Tian, and Ji]{xu2021recu}
Zihan Xu, Mingbao Lin, Jianzhuang Liu, Jie Chen, Ling Shao, Yue Gao, Yonghong Tian, and Rongrong Ji.
\newblock Recu: Reviving the dead weights in binary neural networks.
\newblock In \emph{Proceedings of the IEEE/CVF International Conference on Computer Vision (ICCV)}, pages 5198--5208, 2021.

\bibitem[Yuan et~al.(2022)Yuan, Xue, Chen, Wu, and Sun]{yuan2022ptq4vit}
Zhihang Yuan, Chenhao Xue, Yiqi Chen, Qiang Wu, and Guangyu Sun.
\newblock Ptq4vit: Post-training quantization for vision transformers with twin uniform quantization.
\newblock In \emph{Proceedings of the European Conference on Computer Vision (ECCV)}, pages 191--207. Springer, 2022.

\bibitem[YVINEC et~al.(2023)YVINEC, Dapogny, Cord, and Bailly]{yvinec2022powerquant}
Edouard YVINEC, Arnaud Dapogny, Matthieu Cord, and Kevin Bailly.
\newblock Powerquant: Automorphism search for non-uniform quantization.
\newblock In \emph{Proceedings of the International Conference on Learning Representations (ICLR)}, 2023.

\bibitem[Zhang et~al.(2022)Zhang, Peng, Wu, Liu, Xiao, Fu, and Yuan]{zhang2022minivit}
Jinnian Zhang, Houwen Peng, Kan Wu, Mengchen Liu, Bin Xiao, Jianlong Fu, and Lu Yuan.
\newblock Minivit: Compressing vision transformers with weight multiplexing.
\newblock In \emph{Proceedings of the IEEE/CVF Conference on Computer Vision and Pattern Recognition (CVPR)}, pages 12145--12154, 2022.

\bibitem[Zheng et~al.(2021)Zheng, Lu, Zhao, Zhu, Luo, Wang, Fu, Feng, Xiang, Torr, et~al.]{zheng2021rethinking}
Sixiao Zheng, Jiachen Lu, Hengshuang Zhao, Xiatian Zhu, Zekun Luo, Yabiao Wang, Yanwei Fu, Jianfeng Feng, Tao Xiang, Philip~HS Torr, et~al.
\newblock Rethinking semantic segmentation from a sequence-to-sequence perspective with transformers.
\newblock In \emph{Proceedings of the IEEE/CVF conference on computer vision and pattern recognition (CVPR)}, pages 6881--6890, 2021.

\bibitem[Zhong et~al.(2022{\natexlab{a}})Zhong, Lin, Chen, Li, Shen, Chao, Wu, and Ji]{zhong2022fine}
Yunshan Zhong, Mingbao Lin, Mengzhao Chen, Ke Li, Yunhang Shen, Fei Chao, Yongjian Wu, and Rongrong Ji.
\newblock Fine-grained data distribution alignment for post-training quantization.
\newblock In \emph{Proceedings of the European Conference on Computer Vision (ECCV)}, pages 70--86. Springer, 2022{\natexlab{a}}.

\bibitem[Zhong et~al.(2022{\natexlab{b}})Zhong, Lin, Li, Li, Shen, Chao, Wu, and Ji]{zhong2022dynamic}
Yunshan Zhong, Mingbao Lin, Xunchao Li, Ke Li, Yunhang Shen, Fei Chao, Yongjian Wu, and Rongrong Ji.
\newblock Dynamic dual trainable bounds for ultra-low precision super-resolution networks.
\newblock In \emph{Proceedings of the European Conference on Computer Vision (ECCV)}, pages 1--18. Springer, 2022{\natexlab{b}}.

\bibitem[Zhong et~al.(2022{\natexlab{c}})Zhong, Lin, Nan, Liu, Zhang, Tian, and Ji]{zhong2022intraq}
Yunshan Zhong, Mingbao Lin, Gongrui Nan, Jianzhuang Liu, Baochang Zhang, Yonghong Tian, and Rongrong Ji.
\newblock Intraq: Learning synthetic images with intra-class heterogeneity for zero-shot network quantization.
\newblock In \emph{Proceedings of the IEEE/CVF Conference on Computer Vision and Pattern Recognition (CVPR)}, pages 12339--12348, 2022{\natexlab{c}}.

\bibitem[Zhong et~al.(2022{\natexlab{d}})Zhong, Lin, Zhang, Nan, Chao, and Ji]{zhong2022exploiting}
Yunshan Zhong, Mingbao Lin, Yuxin Zhang, Gongrui Nan, Fei Chao, and Rongrong Ji.
\newblock Exploiting the partly scratch-off lottery ticket for quantization-aware training.
\newblock \emph{arXiv preprint arXiv:2211.08544}, 2022{\natexlab{d}}.

\bibitem[Zhong et~al.(2023{\natexlab{a}})Zhong, Lin, Xie, Zhang, Chao, and Ji]{zhong2023distribution}
Yunshan Zhong, Mingbao Lin, Jingjing Xie, Yuxin Zhang, Fei Chao, and Rongrong Ji.
\newblock Distribution-flexible subset quantization for post-quantizing super-resolution networks.
\newblock \emph{arXiv preprint arXiv:2305.05888}, 2023{\natexlab{a}}.

\bibitem[Zhong et~al.(2023{\natexlab{b}})Zhong, Lin, Zhou, Chen, Zhang, Chao, and Ji]{zhong2023multiquant}
Yunshan Zhong, Mingbao Lin, Yuyao Zhou, Mengzhao Chen, Yuxin Zhang, Fei Chao, and Rongrong Ji.
\newblock Multiquant: A novel multi-branch topology method for arbitrary bit-width network quantization.
\newblock \emph{arXiv preprint arXiv:2305.08117}, 2023{\natexlab{b}}.

\bibitem[Zhu et~al.(2020)Zhu, Su, Lu, Li, Wang, and Dai]{zhu2020deformable}
Xizhou Zhu, Weijie Su, Lewei Lu, Bin Li, Xiaogang Wang, and Jifeng Dai.
\newblock Deformable detr: Deformable transformers for end-to-end object detection.
\newblock \emph{arXiv preprint arXiv:2010.04159}, 2020.

\end{thebibliography}
